\documentclass{article}
\usepackage{spconf,amsmath,graphicx,epstopdf}
\usepackage{verbatim}
\usepackage{subfigure}

\title{Beyond Context: Exploring Semantic Similarity for Tiny Face Detection}
\name{Yue Xi$^{1,2}$, Jiangbin Zheng$^{1,}$*, Xiangjia He$^{2,}$*, Wenjing Jia$^{2}$, Hanhui Li$^{2,3}$\thanks{*Xiangjian He and Jiangbin Zheng are the co-corresponding authors
		for this paper.}}
\address{$^{1}$School of Computer Science, Northwestern Polytechnical University, P.R.China \\ 
$^{2}$Global Big Data Technologies Centre (GBDTC), University of Technology Sydney, Australia \\
$^{3}$ School of Data and Computer Science, Sun Yat-sen University, P.R.China}
\begin{document}
%
\maketitle

\begin{abstract}

%
%
%
Tiny face detection aims to find faces with high degrees of variability in scale, resolution and occlusion in cluttered scenes. Due to the very little information available on tiny faces, it is not sufficient to detect them merely based on the information presented inside the tiny bounding boxes or their context. In this paper, we propose to exploit the semantic similarity among all predicted targets in each image to boost current face detectors. To this end, we present a novel framework to model semantic similarity as pairwise constraints within the metric learning scheme, and then refine our predictions with the semantic similarity by utilizing the graph cut techniques. Experiments conducted on three widely-used benchmark datasets have demonstrated the improvement over the-state-of-the-arts gained by applying this idea. 

\end{abstract}
\section{Introduction}
\label{sec:intro}

\begin{figure}[t]
	
	\begin{minipage}[t]{1.0\linewidth}
		\centering
		\centerline{\includegraphics[width=0.75\textwidth]{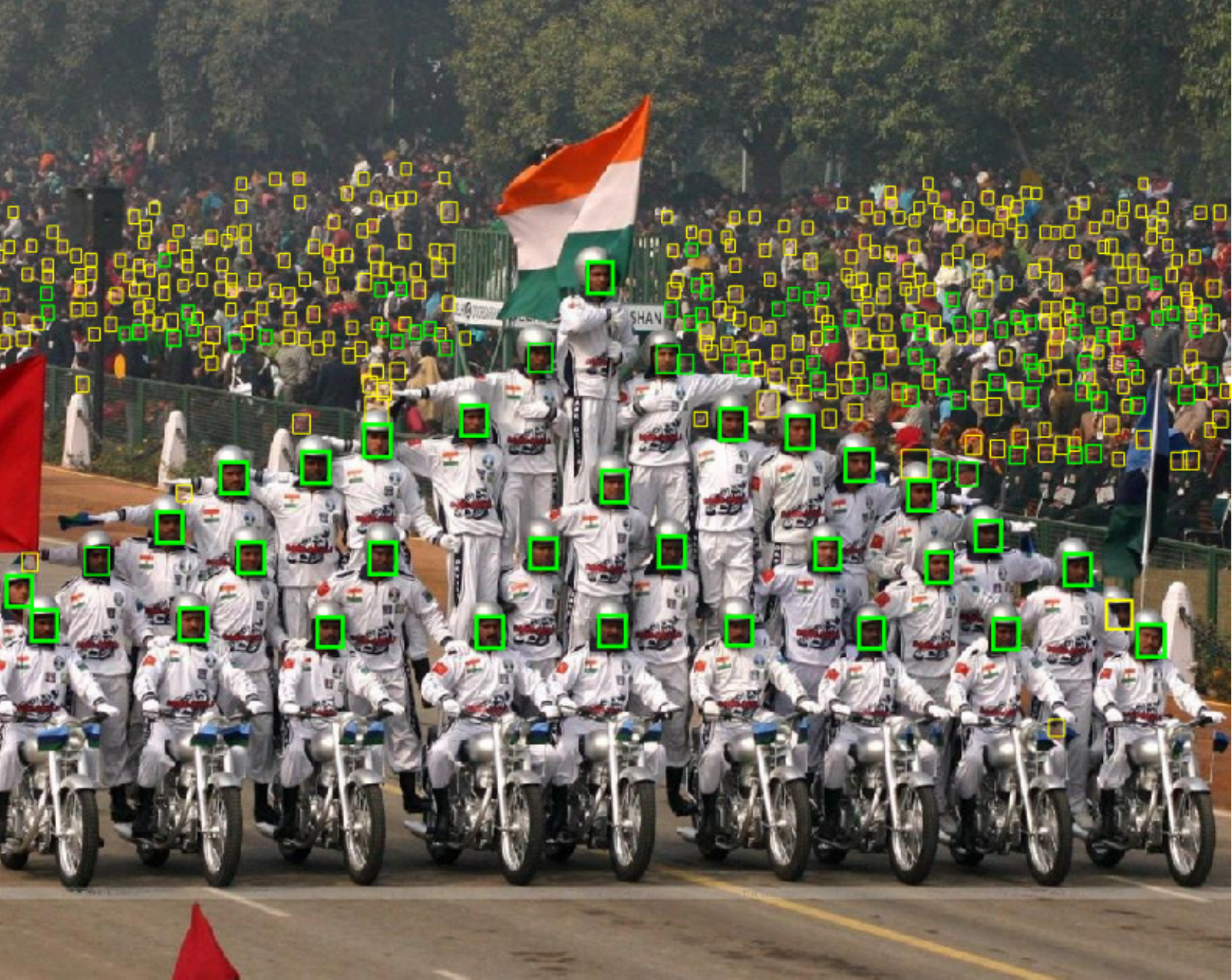}}
		\caption{Tiny faces detected with our proposed approach (shown as yellow and green boxes) and the HR approach~\cite{hu2017finding} (shown as green boxes).}
		\label{fig:tiny face images}
	\end{minipage}
	
\end{figure}

Robust face detection is one of the ultimate components to support various facial 
related problems, such as face alignment \cite{xiong2013supervised}\cite{zhu2016face}, face recognition \cite{parkhi2015deep}\cite{schroff2015facenet}\cite{zhu2015high}, face verification \cite{sun2014deep}, and face tracking \cite{kim2008face}, etc. From the cornerstone by Viola-Jones \cite{Viola2004Robust} to the recent work by Hu et al. \cite{hu2017finding}, the performance of face detection has been 
improved dramatically. 
The recent introduction of the WIDER
face dataset \cite{yang2016wider}, which contains a large number of small faces, exposes the performance gap between humans and 
the current face detection techniques due to a number of challenges in practice. 
Different from the classical face detection, tiny face detection mainly focuses on low-resolution, large scale variation and serious occlusion, as shown in Fig.~\ref{fig:tiny face images}. 
All of these challenges suggest the information on small objects is far too limited.

The existing methods for finding small objects in imageries can be grouped into three categories. The first group (e.g., \cite{liu2016ssd}) aims to extract scale-invariant features using pre-trained deep networks. However, their performance drops dramatically as the target faces become too small. Another group tries to generate additional information inside the objects by interpolation. For example, the work in~\cite{hu2017finding} demonstrated that interpolating the lowest layer of image pyramid was significantly beneficial for capturing small objects. 
The last group (e.g., \cite{Le2015ASW}) seeks to incorporate information surrounding the objects (i.e., context) in order to improve the performance of tiny face detection. It is clear that computer vision needs additional contextual information to accurately classify small faces. Is there another way to improve the performance of small object detection?

\begin{figure*}[t]
	\centering
	\begin{minipage}[t]{0.98\linewidth}
		\centering
		\centerline{\includegraphics[width=0.95\textwidth]{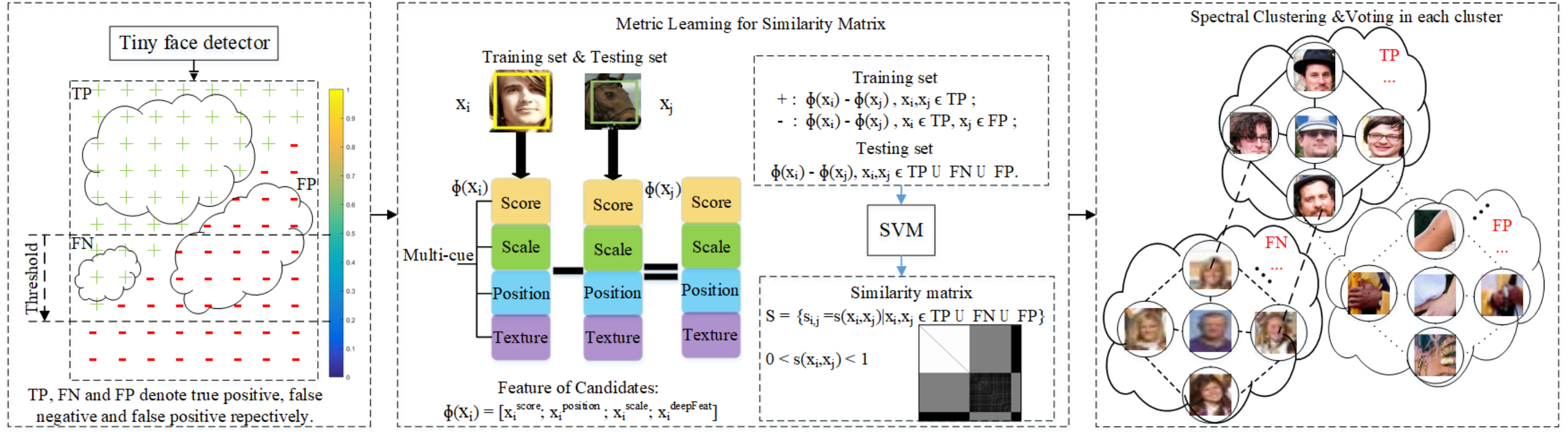}}
		\caption {The framework of our proposed MLGC for high-density tiny face detection. }
		\label{fig 3 frame}
	\end{minipage}
	
\end{figure*}

Note that, the existing classification-based tiny face detectors simply apply a 
threshold on a classification score to determine whether the corresponding candidate is face or non-face, as shown in the first stage of Fig.~\ref{fig 3 frame}. 
However, the optimal threshold is often difficult to obtain.
In this paper, 
we propose a novel idea to exploit the semantic information (consisting of spatial locations, scales and textures) of a candidate's neighbors to classify a target to face or background.
Specifically, based on such semantic information, we try to group all of the faces into one cluster, while backgrounds are kept far away from the cluster. 
For this purpose, we propose a Metric Learning and Graph-Cut (MLGC) framework, which carries out further classification on the candidates produced by other object detectors. 
Fig.~\ref{fig 3 frame} illustrates the framework of this idea.  

We first obtain a high-recall classifier which aims to retrieve all of the targets in an image, but may unavoidably introduce lots of false positives. 
Our focus is to retrieve faces with low classification scores but remove these false positives. 
In order to do this, we design a metric learning method to learn a similarity matrix to evaluate the similarity of each pair of candidates. 
A graph model is built to represent the similarity matrix of these candidates. 
The graph cut technique is utilized to divide the graph into several groups where candidates in the same group are similar and those in different groups are dissimilar to each other. 
Finally, the candidates in each group are classified into faces or non-faces, correspondingly, by voting.

The main contributions of this paper can be highlighted as follows. 
First, aiming to boost the detection performance, we propose a novel metric learning and graph-cut framework to exploit the semantic information between targeting objects' neighbors. 
Secondly, to depict local neighborhood relationships, we introduce a pairwise constraint into the tiny face detector to improve the detection accuracy. 
Thirdly, to realize such a pairwise constraint, we convert the problem of regression that estimates the similarity between different candidates into a classification problem that produces the scores of classification for each pair of candidates.

\section{Related Work}
\label{sec:related}

Face detection is a classic topic in computer vision. 
The pioneer work on the topic was published by Viola and 
Jones \cite{Viola2004Robust} who designed a cascade of weak classifiers using Haar features and AdaBoost for fast and robust face detection. Similar in spirit, numerous approaches have been developed to 
improve the performance with more sophisticated hand-crafted features \cite{yang2014aggregate} and more powerful classifiers \cite{Pham2007Fast}. 
However, these methods using non-robust hand-crafted features
and optimized each components independently, and hence led to sub-optimal face detection results. 
Recently, face detectors based on \emph{CNNs} \cite{li2015convolutional}\cite{yang2015convolutional}\cite{yang2015facial} have greatly bridged the gap between human vision and artificial detectors.

Tiny face detection aims to detect a large number of small faces in crowded and cluttered scenes. 
It is totally different from detecting normal faces, because the cues for detecting a 3-pixel tall face are fundamentally different from those for detecting a 300-pixel tall face \cite{hu2017finding}. Bell~\cite{bell2016inside} presented the Inside-Outside Net (ION) to model the context outside a region of interest and showed improvements on small object detection.
Very recently, Hu and Ramanan \cite{hu2017finding} designed a foveal descriptor that captured both coarse context and high-resolution image features in order to effectively encode context information, which has achieved state-of-the-art performance on the WIDER FACE dataset. 
As we all know, it is not sufficient to detect small objects merely by extracting deep learning features from the texture inside an object region. One main drawback is that, these approaches have neglected local semantic information. We have observed that there exists local coherent relationships in terms of spatial location, scale, and texture in high-density tiny face detection, ignoring the influence of various viewpoints. For example, as shown in Fig.~\ref{fig:tiny face images}, face bounding boxes close to each other are similar in their scales and textures. 
Local semantic information helps tiny face detectors better eliminate false alarms. 
To introduce local coherent relationships, we learn a metric to represent this coherence and use the graph-cut algorithm to divide candidates into several groups, where candidates in the same group are similar, and dissimilar when they are in different groups.
 
\section{The Proposed Method}
\label{sec:metric}

Our goal is to integrate local coherent relationships into tiny face detection. 
In order to represent local coherent relationships, we define pairwise constraints, which are an equivalence
constraint for pairs of data points belonging to same classes, and an inequivalence constraint for pairs of data points belonging to different classes.

As shown in Fig.~\ref{fig 3 frame}, we present a metric learning and graph-cut (MLGC) approach for high-density tiny face detection. 
We first use a linear-SVM to estimate the similarity matrix among all candidates (Sect.~\ref{3.1 Unsupervised Multi-Cues Model}) and then we construct a graph model and use the graph-cut algorithm to divide candidates into several groups (Sect.~\ref{3.2SpecturalClustering}). 
Finally, we design a voting method to classify groups (Sect.~\ref{3.2SpecturalClustering}).

\label{sec:format}

%
%

\subsection{Metric learning based on linear-SVM}
\label{3.1 Unsupervised Multi-Cues Model}

Let $X = \{x_{1}, x_{2}, ..., x_{N}\}$ denote the set of $N$ candidates (i.e., face or non-face bounding boxes). 
To introduce the pairwise constraint,  
we first build a similarity matrix $S = s(x_{i},x_{j}), x_{i}, x_{j} \in X, i,j  = 1,2, ...N$, where $s(x_{i}, x_{j})$ represents the similarity between $x_{i}$ and $x_{j}$. 
$s(x_{i}, x_{j}) = 1$ means that $x_{i}$ has a strong resemblance of $x_{j}$, and $s(x_{i}, x_{j}) = 0$ means that $x_{i}$ is completely different from $x_{j}$.  

In order to obtain the similarity score between two candidates $x_{i}$ and $x_{j}$, we treat it as a classification problem and propose an unsupervised way to obtain the similarity score between two candidates. 
We use SVM to compute the similarity score between two candidates $x_{i}$ and $x_{j}$ based on multiple cues, i.e., the position, scale, classification score and deep features of the candidates, which are concatenated together into a feature vector $\phi(x_{i})$. Note that, classification scores and deep features of a candidate $x_{i}$ are obtained from the tiny face detector \cite{hu2017finding}. During the training stage, we sort $X$ by their scores in descending order. We suppose that $X_{Top}$ denotes the top $10\%$ of $X$ which are face patches, while $X_{Bottom}$ denotes the bottom $10\%$
of the non-face patches in $X$. As shown in Fig.~\ref{fig 3 frame}, in Stage 2 of our MLGC, we build a training set $\{(x_{11}^{'},y_{11}^{'}), (x_{12}^{'},y_{12}^{'}), ... (x_{nn}^{'},y_{nn}^{'})\}$, $x_{ij}^{'} = \phi(x_{i}) - \phi(x_{j})$, $y_{ij}^{'} = \{0,1\}$. If $x_{i}, x_{j} \in X_{Top},  y_{ij}^{'} = 1$. If $x_{i} \in X_{Top}, x_{j} \in X_{Bottom},  y_{ij}^{'} = 0$. 
During the testing stage, we feed $x_{ij}^{'} = \phi(x_{i}) - \phi(x_{j})$ to the SVM classifier, and then use the output score as the similarity score $s(x_{i},x_{j})$ between $x_{i}$ and $x_{j}$. Thus, we build the similarity matrix $S$.

%

\subsection{Graph-cut based on spectral clustering}
\label{3.2SpecturalClustering}

Given a set of candidates $X=\{x_{1},x_{2},...,x_{N}\}$ and a similarity matrix $S$, our goal is to cluster $X$ into different groups. Candidates are similar when they are in the same group, and are dissimilar when they are in different groups. In this work, we adopt the graph-cut algorithm for this purpose. First, we build a graph model $G = (V,E)$ to represent $X$, where each vertex $v_{i}\in V$ represents a candidate $x_{i}$, and $e_{ij}\in E$ represents the similarity $s(x_{i},x_{j})$ between the corresponding candidates $x_{i}$ and $x_{j}$. Then, clustering $X$ into groups can be reformulated with the graph model represented in Eq. \ref{equ:cutWithoutSize}.
We want to find a partition of the graph so that the weights of edges between different subgraphs are very low    
(indicating that points in different clusters are dissimilar from each other) and the weights of edges in the same group are very high (meaning that points within the same cluster are similar to each other). 
Formally, 
\begin{equation}
	cut(A_{1},A_{2},...,A_{k}) = \frac{1}{2} \sum_{i=1}^{k}W(A_{i},\bar{A_{i}})
	\label{equ:cutWithoutSize}
\end{equation}
where  $A_{i}\subset V, A_{i}\cap A_{j} = \emptyset$ and $A_{1}\cup A_{2} \cup ... \cup A_{k} = V$, $W(A_{i},\bar{A_{i}}) = \sum_{m\in {A_{i}},n\in \bar{A_{i}}}w_{mn}$, $w_{mn}=exp({-S_{mn}}/{2\delta^2})$ used to boost local neighborhood relationships. 

However, the solution simply separates one individual vertex from the rest of the graph.
To avoid unbalanced graph-cut situation that there is a large difference in sizes of subgraphs, we introduce the size of 
subgraph $|A|$ which is the number of vertexes in $A$ to ensure the set of subgraph $\{A_{1},A_{2},...,A_{k}\}$ is reasonably large. So, we can transform Eq.~\ref{equ:cutWithoutSize} as follows:

\begin{equation}
	cut(A_{1},A_{2},...,A_{k}) = \frac{1}{2} \sum_{i=1}^{k} \frac{W(A_{i},\bar{A_{i}})}{|A_{i}|}
	\label{equ:cutSize}
\end{equation}

According to \cite{von2007tutorial}, 
\begin{equation}
\begin{split}
	\arg\min cut(A_{1},A_{2},...,A_{k}) = \mathop{\arg\min}_{H} Tr(H^{T}LH) 
\end{split}
	\label{equ:argmin}
\end{equation}
where 
$L \textrm{ is the Laplacian matrix}, H^{T}H = I$, and the indicator 
$$H = \{h_{1},h_{2},...,h_{k}\}$$ with
\begin{equation}
	h_{i,j}=\left\{ \begin{array}{ll}
	\frac{1}{\sqrt{A_{j}}} & \textrm{if $v_{i}\in A_{j}$}\\
	0 & \text{otherwise}\\
	\end{array} \right.
\end{equation}
for $i = 1,2,...,N; j = 1,2,...,k$.

Eq.~\ref{equ:argmin} is the standard form of a trace minimization
problem. According to the Rayleigh-Ritz theorem~\cite{lutkepohl1997handbook}, the solution is given by choosing the matrix $U$ which contains the first $k$ eigenvectors of $L$ and then uses the $k$-means algorithm on $U$. So, we manage to cluster $X$ into 
$k$ groups $\{A_{1},A_{2},...,A_{k}\}$. 
Finally, candidates in each group are classified to face or non-face class using voting.




\section{Experiments}
\label{sec:Experiments}

In this section, we first demonstrate the effectiveness of our proposesd semantic similarity metric and then evaluate the whole model on three widely-used face detection benchmarks, including WIDER FACE~\cite{yang2016wider}, Annotated Faces in the Wild (AFW)~\cite{zhu2012face} and Pascal Faces~\cite{yan2014face}.

%


To demonstrate the effectiveness of our proposed semantic metric (see Subsection~\ref{3.1 Unsupervised Multi-Cues Model}) for similarity measurement, we create positive samples, i.e., the ground truth face regions, and negative samples which are patches randomly sampled from background, and evaluate the discriminative ability of the computed similarity matrix on the WIDER FACE validation set. 
The average precision in each image on the validation set is 
$79.58\%$ in the testing set composed of both positive and negative samples,
$72.25\%$ in the set of positive samples only, 
and $86.75\%$ in the set of negative samples only.


\subsection{The AFW and PASCAL FACE Dataset Results}

The AFW dataset has 205 images containing in total 473 labelled faces. 
We evaluate our model against the HR~\cite{hu2017finding}, DPM \cite{felzenszwalb2010object}, Headhunter, SquaresChnFtrs~\cite{mathias2014face}, Structured Models \cite{yan2014face}, Shen et al.~\cite{shen2013detecting}, TSM~\cite{zhu2012face} and commercial detectors (e.g., Face.com, Face++ and Picasa). 
As illustrated in Fig.~\ref{exp:allb}, our MLGC
outperforms all other detectors on precision-recall (PR) curves.

The PASCAL FACE dataset contains 1,335 labeled faces in 851 images, which are collected from PASCAL person layout subset. Because this paper focuses on face detection, we ignore images without persons from the original dataset, similar like DPM~\cite{felzenszwalb2010object}. 
We also evaluate our model against the HR~\cite{hu2017finding}, DPM~\cite{felzenszwalb2010object}, Headhunter, SquaresChnFtrs~\cite{mathias2014face}, Structured Models~\cite{yan2014face}, Shen et al.~\cite{shen2013detecting}, TSM~\cite{zhu2012face} and commercial detectors (e.g., Face++ and Picasa). 
As shown in Fig.~\ref{exp:allc}, our MLGC
outperforms all other detectors on PR curves.

\subsection{Results Obtained on the WIDER FACE Dataset}

The WIDER FACE Dataset is one of the most challenging public face datasets due to the variety of face scales and occlusion. 
It contains 32,203 images split into training (40\%), validation (10\%) and testing (50\%) set. 
The validation set and testing set are divided into ``easy", ``medium", and ``hard" subsets according to the difficulties of the detection. 

We compare our MLGC with the HR~\cite{hu2017finding}, MSCNN~\cite{cai2016unified}, ScaleFace~\cite{yang2017face}, CMS-RCNN~\cite{zhu2017cms} and Multitask Cascade CNN~\cite{zhang2016joint}. 
The PR curves on the testing set is presented in Fig. ~\ref{exp:alla}, and our method outperforms HR by 0.2\% in ``easy" subset.
The PR curves on the validation set is presented in Fig. \ref{fig:overall} and our method outperforms the HR by 0.5\%, 0.2\%, 0.3\%, in ``easy", ``medium" and ``hard" subsets respectively.

\section{Conclusion}
\label{sec:conclu}

In this paper, aiming to improve the performance of tiny face detection, we have proposed a novel idea to exploit the semantic similarity between targeting objects' neighbors and created a pairwise constraint to depict such semantic similarity. 
Then, a framework which adopts the metric learning and graph-cut techniques has been formulated to boost the accuracy of existing tiny object classifiers. 
Experiments conducted on three widely-used benchmark datasets for face detection have demonstrated the improvement over the state-of-the-arts by applying this idea. 
The mechanism of our proposed framework is generic indicating that the framework has a great potential being applied on other small and generic object detectors. 
This forms our work for the next step. 


\bibliographystyle{IEEEbib}
\bibliography{mybibtex}

\end{document}